\documentclass[sigplan,screen,nonacm]{acmart}

\usepackage{algorithm}
\usepackage{algpseudocode}
\usepackage{booktabs}
\usepackage{pifont}

\newcommand{\name}{\textsc{llada.cpp}}

\begin{document}

\title{Efficient On-Device Diffusion LLM Inference \\ with Mobile NPU}

\author{Tuowei Wang}
\authornote{These authors contributed equally to this work.}
\affiliation{
  \institution{Tsinghua University}
  \country{China}
}

\author{Yanfan Sun}
\authornotemark[1]
\affiliation{
  \institution{Beihang University}
  \country{China}
}

\author{Ju Ren}
\authornote{Corresponding author: renju@tsinghua.edu.cn}
\affiliation{
  \institution{Tsinghua University}
  \country{China}
}

\begin{abstract}
Diffusion large language models (dLLMs) accelerate generation by denoising multiple tokens in parallel, making them attractive for latency-sensitive mobile inference. However, repeated denoising introduces substantial computation on smartphones. Mobile neural processing units (NPUs) offer high-throughput dense matrix computation, but efficiently exploiting them remains challenging: token commitment shrinks per-block effective workloads, token revision complicates KV cache reuse, and limited NPU-visible address space incurs costly remapping and data transfer overheads.

In this paper, we propose \name{}, the first NPU-aware inference framework for accelerating dLLMs on smartphones. \name{} aligns block-wise dLLM inference with the execution characteristics of mobile NPUs through three techniques. (1) \textit{Multi-Block Speculative Decoding} fills the shrinking workload in late-stage current-block decoding with speculative future-block tokens. (2) \textit{Dual-Path Progressive Revision} keeps committed tokens revisable until stable and refreshes unstable tokens through a CPU-side path without stalling dense NPU execution. (3) \textit{Swap-Optimized Memory Runtime} compacts NPU-visible address layouts and overlaps data staging with NPU computation to reduce remapping and transfer overheads. We implement \name{} as an end-to-end framework and evaluate it across diverse hardware platforms and dLLM workloads. \name{} reduces LLaDA-8B generation latency by 17$\times$-42$\times$ over the CPU baseline with prefix KV cache reuse, while preserving generation quality.
\end{abstract}

\maketitle
\pagestyle{plain}

\section{Introduction}
Large language models (LLMs) are becoming a core interface for mobile computing, enabling applications such as personalized assistants~\cite{mobile-llm,mobile-ui}, multimodal interaction~\cite{mobile-vlm,app-agent}, and context-aware automation~\cite{autodroid,mobile-gpt}. Despite their success, LLM inference remains dominated by autoregressive decoding~\cite{attention}, which generates outputs token by token and therefore incurs latency that scales with output length. On mobile devices, this latency pressure is exacerbated by limited memory bandwidth and strict power budgets, impeding the development of interactive LLM applications.

Recent \textit{diffusion large language models (dLLMs)}~\cite{llada,d3pm,mercury} offer a promising alternative. Instead of generating tokens one by one, dLLMs iteratively denoise a masked sequence and update multiple token positions in parallel. To make this iterative process practical for long-sequence serving, existing systems often adopt \textit{block-wise decoding}~\cite{block-diffusion}, where the output sequence is partitioned into consecutive blocks (e.g., 32 tokens). Blocks are decoded sequentially from left to right to enable prefix key-value (KV) cache reuse, while tokens within each block are updated in parallel through iterative denoising. This non-autoregressive formulation exposes sequence-level parallelism and shifts the latency bottleneck from output length to the number of denoising steps.

\begin{figure}[t]
    \centering
    \includegraphics[width=1.0\linewidth]{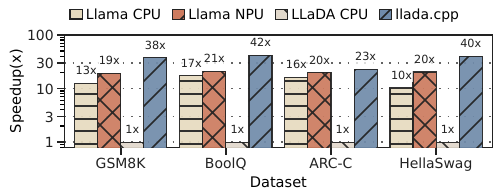}
    \caption{End-to-end generation speedup for 128-token outputs on the OnePlus Ace5 Pro with the SM8750 SoC.}
    \label{fig:core-result}
    \vspace{-0.5cm}
\end{figure}

Although dLLMs reduce token-level sequential dependencies, they introduce a new performance bottleneck: \textit{repeated parallel computation}. A dLLM typically performs multiple denoising iterations, each requiring expensive transformer computation over the full sequence. Moreover, because dLLM iterations repeatedly revise token states, reusing the KV cache across iterations within a block becomes difficult. On resource-constrained mobile devices such as smartphones, this repeated parallel computation can impose substantial computational burden. As shown in Figure~\ref{fig:core-result}, such overhead can offset the benefits of dLLM over autoregressive decoding.

In this paper, we identify a key architectural opportunity: the parallel decoding nature of dLLMs aligns well with the tensor-computation capability of modern mobile \textit{neural processing units (NPUs)}~\cite{hexagon-npu,amd-npu}. Mobile NPUs are designed to execute dense tensor operations with high throughput and energy efficiency. For example, Qualcomm reports that the Hexagon NPU in Snapdragon X Elite delivers up to 45 TOPS INT8 performance~\cite{snapdragon-x-elite}. While autoregressive decoding exposes limited parallelism at each generation step, block-wise dLLM inference naturally creates large parallel workloads within each block and across denoising iterations. This observation creates a new opportunity for on-device acceleration: \textit{by rethinking dLLM inference around the execution characteristics of mobile NPUs}, we can convert the repeated parallel computation of dLLMs from a liability into an advantage.

However, efficiently realizing dLLM inference on smartphones requires more than naive NPU offloading. Although mobile NPUs are well suited for dense tensor operations, they favor static, regular, and shape-stable execution, whereas block-wise dLLM inference is inherently dynamic. This mismatch manifests at three levels: (1) At the \textbf{token-commit level}, as denoising leaves fewer masked tokens in each block, the effective workload shrinks, making late-stage NPU forwards poorly amortized. (2) At the \textbf{token-revision level}, committed tokens may still require revision as more context becomes available, introducing sparse and irregular KV cache updates. (3) At the \textbf{memory-access level}, the working set of dLLM inference often exceeds the limited NPU-visible address space, forcing frequent data swaps with system memory and incurring costly mapping and transfer overheads.

To address these challenges, we propose \name{}, the first NPU-aware inference framework for accelerating dLLMs on smartphones. The key idea of \name{} is to \textit{align the parallel decoding nature of dLLMs with the powerful tensor computation capabilities of mobile NPUs}, translating improved hardware utilization into lower generation latency. Specifically, \name{} introduces three key components:

\noindent(1) \textbf{Multi-Block Speculative Decoding} improves NPU utilization during the later stages of denoising. \name{} speculatively incorporates tokens from future blocks to maintain sufficient parallel workload, while preserving the original commitment order through block-wise acceptance.

\noindent(2) \textbf{Dual-Path Progressive Revision} enables efficient token revision without disrupting dense NPU execution. \name{} uses a fine-grained algorithm to identify tokens that require revision and offloads the corresponding updates and logits computation to specialized CPU-side kernels.

\noindent(3) \textbf{Swap-Optimized Memory Runtime} targets the memory-access bottleneck caused by the limited NPU-visible address space. \name{} uses graph-level lifetime and access-order information to build compact VA layouts, reducing fragmentation and remapping overhead, and pipelines data staging with NPU execution to hide data-movement latency.

We implement \name{} as an end-to-end framework and evaluate it on diverse hardware platforms and dLLM workloads. \name{} reduces LLaDA-8B end-to-end latency by 17$\times$-42$\times$ over the CPU baseline with prefix KV cache reuse, while preserving generation quality. Notably, \name{} achieves up to 3.9$\times$ speedup for LLaDA-8B over an equal-size autoregressive model, highlighting the advantage of diffusion parallel generation on mobile devices.

In summary, this paper makes the following contributions:
\begin{itemize}
    \item We identify the mismatch between static mobile NPU execution and dynamic block-wise dLLM inference, and characterize three key challenges in token commitment, token revision, and memory access.
    \item We propose \name{}, the first NPU-aware inference framework for accelerating dLLMs on smartphones, combining algorithmic and system-level optimizations across the NPU, CPU, and system memory.
    \item We evaluate \name{} across diverse hardware platforms and dLLM workloads, showing that it effectively translates dLLM parallelism into lower smartphone latency than autoregressive and CPU-only baselines.
\end{itemize}

\section{Background and Motivation}
\subsection{Diffusion Large Language Model}
Autoregressive large language models (LLMs) generate text in a strictly left-to-right manner, where each decoding step predicts only the next token conditioned on previously generated tokens. This paradigm has become the dominant generation approach in current LLMs and naturally supports incremental decoding with KV cache reuse. However, its sequential dependency exposes limited token-level parallelism. As a result, even with token-level optimizations, generation latency remains fundamentally tied to the output length.

\begin{figure}[t]
    \centering
    \includegraphics[width=1.0\linewidth]{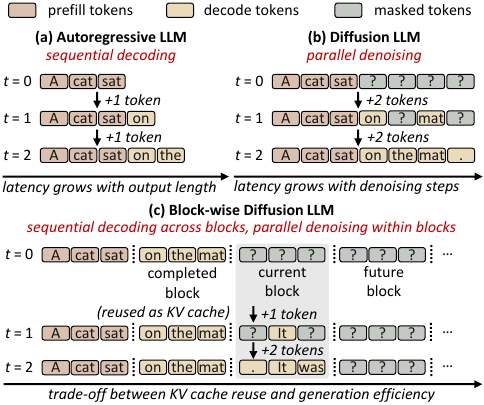}
    \caption{Comparison of decoding paradigms: (a) autoregressive, (b) diffusion, and (c) block-wise diffusion LLM decoding.}
    \label{fig:paradigm}
    \vspace{-0.5cm}
\end{figure}

Diffusion large language models (dLLMs) provide an alternative to autoregressive generation. As shown in Figure~\ref{fig:paradigm}(b), they generate text by iteratively denoising masked token sequences rather than producing tokens one at a time. At each step, the model predicts unresolved positions in parallel and updates a confidence-selected subset, gradually refining the sequence into the final output. This shifts latency from output length to the number of denoising steps, while converting token-wise decoding into sequence-wise matrix workloads that better utilize accelerators and reduce generation overhead. These properties make dLLMs particularly attractive for latency-sensitive scenarios on mobile devices.

\noindent\textbf{Block-Wise Decoding.} A key limitation of vanilla dLLM decoding is the lack of effective KV cache reuse. Since the model repeatedly denoises a sequence whose token states may change across iterations, cached KV states can quickly become stale, forcing the runtime to recompute attention states over a large portion of the sequence. Block-wise decoding addresses this issue and makes dLLM generation more practical for long sequences. As illustrated in Figure~\ref{fig:paradigm}(c), instead of repeatedly denoising the entire growing sequence, the output is divided into consecutive blocks. Blocks are decoded from left to right, while tokens within the current block are denoised in parallel over multiple steps. This structure localizes repeated computation and enables KV cache reuse across block boundaries. Therefore, block-wise decoding retains intra-block diffusion parallelism while imposing a structured left-to-right workflow across blocks.

\noindent\textbf{KV Cache Refresh.} While block-wise decoding enables KV cache reuse across completed blocks, KV states within the current block remain inherently dynamic. Unlike autoregressive decoding, where generated tokens become fixed once appended to the prefix, dLLM decoding repeatedly refines masked tokens over multiple denoising steps. As token identities change, the corresponding KV states computed in earlier steps may become stale. Table~\ref{tab:kv-stale} shows that reusing such stale KV cache entries can make attention operate on outdated representations, causing the model to condition on inconsistent context and degrading generation quality. Therefore, determining when and how to refresh KV cache becomes a key design problem for practical dLLM systems.

\begin{table}[t]
    \centering
    \caption{Accuracy impact of block-wise decoding. The baseline follows vanilla dLLM decoding, which repeatedly denoises the entire growing sequence, while block-wise decoding reuses the prefix KV cache from completed blocks.}
    \label{tab:kv-stale}
    \small
    \setlength{\tabcolsep}{3pt}
    \begin{tabular*}{\columnwidth}{l@{\extracolsep{\fill}}ccc}
    \toprule
    Dataset & Vanilla dLLM decoding & Block-Wise Decoding & $\Delta$ \\
    \midrule
    GSM8K & 78 / 200 (39.0\%) & 65 / 200 (32.5\%)  & -6.5 \\
    BoolQ & 165 / 200 (82.5\%) & 157 / 200 (78.5\%) & -4.0 \\
    ARC-C & 168 / 200 (84.0\%) & 161 / 200 (80.5\%)  & -3.5 \\
    HellaSwag & 102 / 200 (51.0\%) & 93 / 200 (46.5\%)  & -4.5 \\

    \bottomrule
    \end{tabular*}
\end{table}

\begin{table}[t]
    \centering
    \caption{Latency of one LLaDA-8B denoising step pass with 32-128 active tokens on the CPU and NPU.}
    \label{tab:npu-dense-opportunity}
    \small
    \setlength{\tabcolsep}{3pt}
    \begin{tabular*}{\columnwidth}{r@{\extracolsep{\fill}}rrr}
    \toprule
    Token work & CPU latency & NPU latency & Speedup \\
    \midrule
    32  & 4239.6~ms  & 478.1~ms  & 8.87$\times$ \\
    64  & 7728.4~ms  & 861.6~ms  & 8.97$\times$ \\
    128 & 14758.7~ms & 1531.6~ms & 9.64$\times$ \\
    \bottomrule
    \end{tabular*}
\end{table}

\subsection{Opportunity: Mobile Neural Processing Unit}\label{sec:npu}
Modern mobile devices increasingly integrate neural processing units (NPUs) as first-class compute engines for on-device deep learning inference. Unlike CPUs, which provide lightweight and flexible computation for control-intensive tasks, and GPUs, which are primarily optimized for graphics rendering and often incur high-cost synchronization, NPUs employ dedicated datapaths to execute dense tensor operations efficiently~\cite{heterollm}. As shown in Table~\ref{tab:npu-dense-opportunity}, NPUs exhibit clear performance advantages over CPUs on regular LLaDA denoising workloads such as matrix multiplication.

\noindent\textbf{Key Insight.} In this work, we observe that block-wise dLLM decoding naturally exposes a computation unit that aligns well with the execution model of mobile NPUs. Autoregressive decoding typically advances only one token per step, leading to small matrix shapes and low NPU utilization. In contrast, a dLLM block contains multiple token positions that are denoised together and can be viewed as a small token batch. This allows block-wise denoising to be mapped to dense matrix workloads on the NPU. Since mobile NPUs are often idle during conventional LLM inference, such mapping enables dLLM generation to exploit near-free additional compute capacity, improving generation efficiency while largely hiding the extra computation of parallel denoising.

\begin{figure}[t]
    \centering
    \includegraphics[width=1.0\linewidth]{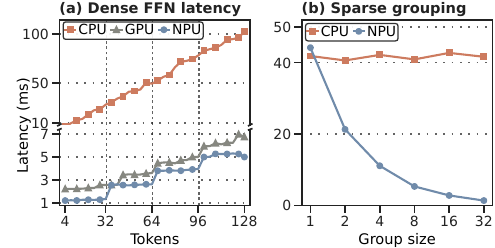}
    \caption{Single FFN down projection matrix multiplication latency on the SM8750 SoC. (a) Measured latency with 4 to 128 tokens on the CPU, GPU, and NPU. (b) Measured latency for processing the same 32 active tokens as sparse groups of different sizes on the CPU and NPU.}
    \label{fig:npu-compute}
    \vspace{-0.5cm}
\end{figure}

Despite this opportunity, effectively exploiting mobile NPUs requires understanding their computation and memory characteristics. Through comprehensive evaluations, our analysis identifies the following four key properties:

\noindent\textbf{(1) Stage Matrix Performance.} On the computation side, mobile NPUs derive most of their throughput from fixed-size tiled matrix engines, such as systolic arrays, which operate on hardware-preferred tile shapes. When tensor dimensions are smaller than the tile size or not divisible by it, the NPU compiler inserts internal padding to form aligned tiles, and these padded elements still consume matrix-engine cycles. Consequently, matrix-multiplication latency exhibits a stage-like pattern rather than scaling smoothly with tensor size. As shown in Figure~\ref{fig:npu-compute}(a), latency remains nearly constant within the same tile-aligned range, but increases abruptly once the tensor shape crosses a tile boundary.

\noindent\textbf{(2) Weak Vector Support.} Beyond tiled matrix computation, mobile NPUs provide much weaker support for irregular or vector-style operations. While NPUs excel at matrix multiplication through dedicated matrix engines, their general-purpose vector units typically offer much lower compute throughput and memory bandwidth. As shown in Figure~\ref{fig:npu-compute}(b), vector-style and irregular operations achieve much lower efficiency on the NPU than dense tiled matrix kernels. This motivates a computation split that offloads dense matrix operations to the NPU while leaving irregular control and vector-heavy operations to the CPU.

\begin{figure}[t]
    \centering
    \includegraphics[width=1.0\linewidth]{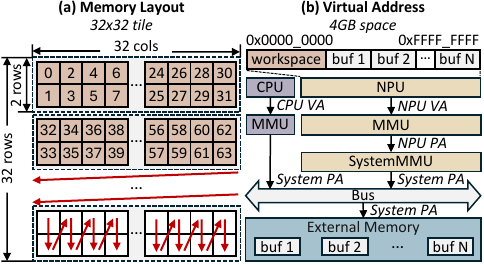}
    \caption{NPU memory characteristics: (a) FP16 HMX tile layout and (b) CPU-NPU system with shared memory.}
    \label{fig:npu-memory}
    \vspace{-0.5cm}
\end{figure}

\noindent\textbf{(3) Specialized Memory Layout.} On the memory side, mobile NPUs rely on specialized data layouts to sustain high matrix throughput. Their matrix engines consume activations and weights in hardware-specific tiled formats. As shown in Figure~\ref{fig:npu-memory}, the FP16 path uses 32$\times$32 matrix tiles with hardware-expected internal ordering, including tile-level inner-product order and row shuffling. Since CPU-oriented tensor layouts are incompatible with the formats directly consumed by NPU matrix engines, additional layout conversion is required, introducing non-trivial runtime overhead.

\noindent\textbf{(4) Limited Virtual Address.} In addition to layout constraints, mobile NPUs access shared buffers with the CPU through a constrained virtual memory mechanism. Before a buffer can be consumed by an NPU kernel, it must be registered and mapped into the NPU virtual address space. This address space is limited by session-level address windows, mapper entries, alignment requirements, and fragmentation. As a result, frequent buffer allocation or remapping can introduce noticeable latency and may even cause mapping failures under memory pressure. Efficient NPU-backed execution therefore requires stable buffer reuse and careful management of the limited NPU virtual address space.

\subsection{Challenges: Static NPUs Meet Dynamic dLLMs}\label{sec:challenge}
Despite the natural alignment between parallel decoding of dLLMs and the tensor computation capability of mobile NPUs, efficient dLLM inference requires more than simply mapping parallel tensor computation onto the hardware. The core challenge of \name{} lies in the mismatch between the \textit{static} execution model of mobile NPUs and the \textit{dynamic} behavior of dLLM inference. Mobile NPUs are optimized for regular, dense, and shape-stable tensor computation, whose efficiency relies on fixed tile granularity, contiguous data layouts, and predictable memory access. In contrast, block-wise dLLM inference continuously changes the active token set, reusable KV states, and accessed data regions across denoising steps. This mismatch manifests in three dimensions:

\begin{figure}[t]
    \centering
    \includegraphics[width=1.0\linewidth]{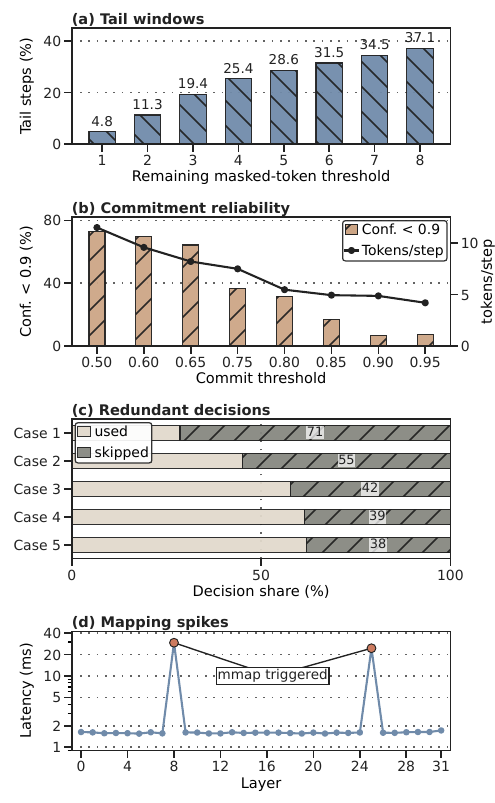}
    \caption{Dynamic behavior of block-wise dLLM inference on mobile NPUs. (a) Share of denoising steps in which the current block has only a small number of remaining masked tokens. (b) For different commit thresholds, the number of tokens committed per step and the fraction of committed tokens whose confidence remains below 0.9 after all tokens in the block are committed. (c) Share of output decisions that are still evaluated and the share that can be skipped after positions become stable. (d) Per-layer FFN operator latency, showing spikes when buffer mappings are triggered to make weights visible in the NPU address space.}
    \label{fig:challenge}
\end{figure}

\noindent(1) \textbf{Token-Commit Level.} In block-wise decoding, tokens in the current block are progressively committed as denoising proceeds. Early in decoding, many masked tokens remain active, providing sufficient parallel workload for each NPU forward pass. Later, however, the number of remaining masked tokens decreases, reducing the effective computation per forward pass and leaving the NPU underutilized. Figure~\ref{fig:challenge}(a) shows that this low-workload tail appears across many denoising steps, exposing little useful token work while still occupying NPU execution slots.

\noindent(2) \textbf{Token-Revision Level.} Block-wise decoding must decide when a predicted token can be exposed to later computation, but an early decision made under incomplete intra-block context is not necessarily reliable. Figure~\ref{fig:challenge}(b) quantifies this tension with a single commit threshold: lowering the threshold commits more tokens per step, improving decoding progress, but after the whole block is committed, a larger fraction of those tokens still have confidence below 0.9. This means that treating every committed token as final prefix context can carry uncertain token states into later computation; the runtime should keep such low-confidence committed tokens revisable. At the same time, revision should not force the system to repeatedly recompute output decisions for positions that are already reliable. Figure~\ref{fig:challenge}(c) shows that many such decisions can be skipped, motivating a revision path that tracks uncertain tokens while avoiding redundant output projection and decision logic for stable positions.

\noindent(3) \textbf{Memory-Access Level.} Mobile NPUs expose limited device-visible address space, which must accommodate model weights, the KV cache, activations, and temporary workspaces. dLLM inference repeatedly accesses and swaps among these data across denoising steps, amplifying the overhead of buffer mapping, data transfer, and replacement. As shown in Figure~\ref{fig:challenge}(d), a representative FFN operator becomes a latency spike when a buffer mapping is triggered, even though the same operator is short when the required buffers are already visible. Across one forward pass, only a small fraction of NPU operators trigger mapping, but their cumulative mapping time is large enough to enter the critical path. Such recurring memory-access overheads can substantially offset the latency reduction from NPU acceleration.

\section{System Design}
\subsection{Overview of \name{}}
\begin{figure}[t]
    \centering
    \includegraphics[width=1.0\linewidth]{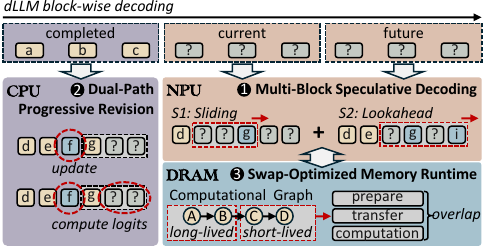}
    \caption{Overview of \name{}.}
    \label{fig:overview}
\end{figure}

We propose \name{}, the first NPU-aware inference framework for dLLMs on smartphones. \name{} follows a block-wise decoding workflow: it partitions the sequence into consecutive blocks and decodes them from left to right. Completed blocks are reused through the KV cache, while tokens in the current block are updated in parallel across denoising steps. Figure~\ref{fig:overview} presents the system overview.

As analyzed in Section~\ref{sec:challenge}, efficient dLLM inference cannot be achieved by simply mapping tensor computation onto NPUs. To address the challenges, \name{} designs three key components: \ding{182} \textbf{Multi-Block Speculative Decoding} addresses the dynamic \textit{token-commit} process within the current block: as the block approaches completion and the number of remaining masked tokens decreases, \name{} speculatively includes tokens from future blocks into the same NPU forward pass to maintain high hardware utilization. \ding{183} \textbf{Dual-Path Progressive Revision} handles the dynamic \textit{token-revision} process in the prefix context: it first identifies revisable tokens through fine-grained state tracking and then offloads their sparse KV cache refreshes and logits computation to a CPU-side path, keeping the NPU dedicated to dense denoising. \ding{184} \textbf{Swap-Optimized Memory Runtime} targets the dynamic \textit{memory-access} process within the limited NPU address space: It uses graph-guided buffer mapping to compact fragmented tensors and reduce avoidable VA swaps, while pipelining mapping preparation, data transfer, and NPU execution to hide the latency of unavoidable swaps.

\subsection{Multi-Block Speculative Decoding}
\begin{figure}[t]
    \centering
    \includegraphics[width=1.0\linewidth]{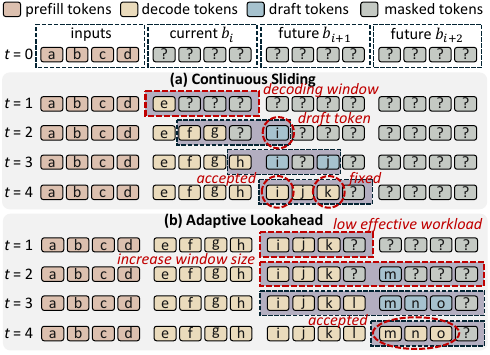}
    \caption{Multi-block speculative decoding. \name{} uses two complementary strategies to expand decoding window.}
    \label{fig:method-decoding}
\end{figure}

In standard block-wise decoding, although each forward pass performs global attention over the block, the number of masked tokens gradually decreases as denoising progresses. As a result, later decoding steps provide insufficient effective workload to fully utilize the NPU compute capacity. As depicted in Figure~\ref{fig:method-decoding}, \name{} dynamically incorporates a subset of tokens from subsequent blocks into current block decoding. By using future tokens as \textit{speculative} work, this design increases the effective workload of each NPU forward pass while preserving the original decoding semantics.

\name{} introduces a \textit{decoding window} as the basic execution unit, defined as the token range included in each NPU forward pass. Unlike a block, which serves as the semantic unit for left-to-right commitment, a decoding window may extend beyond the current block to include tokens from future blocks. These future-block tokens are marked as \emph{draft}: they record early denoising progress but neither update the prefix KV cache nor affect the commitment order. When a future block later becomes current, \name{} re-evaluates its draft tokens and accepts them only if they satisfy the commitment criterion. By decoupling execution from commitment, \name{} adapts the executed token range while preserving decoding semantics. Specifically, \name{} adapts the decoding window through two complementary strategies:

\begin{figure}[t]
    \centering
    \includegraphics[width=1.0\linewidth]{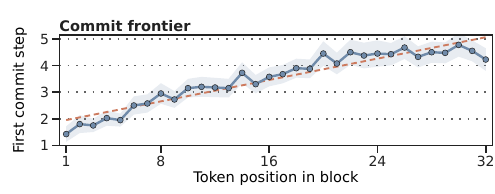}
    \caption{Average first commitment step by token position across 32-token blocks in LLaDA-8B on GSM8K.}
\label{fig:continuous-sliding}
\end{figure}

\noindent\textbf{Strategy \#1: Continuous Sliding.} \name{} first exploits a key property of block-wise diffusion decoding: token commitment within a block follows a left-to-right frontier rather than completing uniformly across all positions. As shown in Figure~\ref{fig:continuous-sliding}, positions closer to the confirmed prefix tend to commit earlier, while later positions require more denoising because they depend on still-uncertain preceding context. This means the useful active region of a block naturally moves rightward as decoding proceeds, creating an opportunity to move the execution window with the commit frontier.

Based on this observation, \name{} slides the decoding window along the token-commit frontier to focus computation on tokens that remain actively denoised. At the beginning of decoding, the window covers the current block. After each denoising step, \name{} commits tokens in the window that satisfy the confidence criterion. Once tokens on the left side of the window are committed, the window slides to the right while keeping its size unchanged, gradually extending into the next block. These future positions then participate in subsequent denoising forward passes and produce tentative predictions and confidence states.

\noindent\textbf{Strategy \#2: Adaptive Lookahead.} Continuous sliding advances future blocks only as committed tokens release space in the window. To better exploit late-stage compute capacity, \name{} introduces adaptive lookahead, which proactively extends the window to include additional future-block tokens in the same NPU forward pass. This strategy is motivated by the non-linear latency behavior of mobile NPUs. As discussed in Section~\ref{sec:npu}, NPU forward latency grows stepwise rather than linearly with the number of participating tokens. Particularly, larger token batches better amortize fixed overheads from data transfers, kernel launches, and buffer allocation, allowing extra future-block tokens to improve NPU utilization with limited marginal latency.

Based on this observation, \name{} converts otherwise wasted compute capacity in late-stage decoding into useful speculative work for future blocks. When only a few masked tokens remain in the current block, each forward pass still pays fixed scheduling, data movement, and kernel execution costs, but exposes too little effective parallel work. \name{} therefore expands the decoding window to include the next block and, when resources allow, additional future blocks in the same forward pass. These future-block tokens reuse the already paid NPU execution overhead for early denoising, increasing the effective workload per forward pass and making the available computation more productive.

\name{} uses an adaptive policy to determine when to expand the decoding window and how many future tokens to include. At each denoising step, it counts the remaining masked tokens in the current block. When this count drops below a threshold, the block is considered to have entered the late decoding stage, triggering lookahead. \name{} then greedily appends future-block tokens according to an offline-profiled NPU latency table, which captures the stepwise NPU latency buckets. This allows the runtime to estimate how many extra tokens can be added before crossing into the next bucket and incurring noticeable marginal latency. A candidate expansion is accepted only if it remains within the profiled latency allowance and its additional memory usage fits the available budget. Otherwise, expansion stops, and the resulting window is used for the next forward pass.

\subsection{Dual-Path Progressive Revision}
Block-wise decoding enables dLLMs to generate tokens in parallel within each block and reuse previous blocks through the prefix KV cache. However, a token reused as a prefix context may still become suboptimal as subsequent tokens are denoised and completed. Conversely, repeatedly denoising such tokens within the current block can be wasteful, as insufficient context limits meaningful refinement. To address this issue, \name{} introduces a staged token stabilization mechanism that identifies tokens requiring revision and applies the corresponding updates through a dedicated CPU-side path, allowing the NPU to focus on dense denoising.

\begin{figure}[t]
    \centering
    \includegraphics[width=1.0\linewidth]{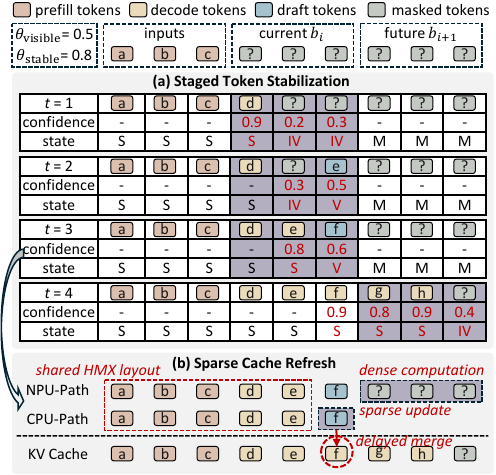}
    \caption{(a) Example of the staged token stabilization algorithm. (b) \name{} offloads unstable-token revision to a CPU path and merges the revised cache in a delayed manner.}
    \label{fig:method-revision}
\end{figure}

\noindent\textbf{Staged Token Stabilization.} \name{} first introduces a three-state transition mechanism to manage the lifecycle of each generated token. The key insight is that a token becoming available for subsequent denoising does not necessarily make it reliable enough to be permanently merged into the prefix KV cache. As shown in Figure~\ref{fig:method-revision}(a), \name{} therefore separates short-term visibility from long-term stability.

Specifically, each generated position is assigned one of three logical states: \textit{invisible (IV)}, \textit{visible (V)}, or \textit{stable (S)}. An \textit{invisible} token is either masked or has a candidate prediction with insufficient confidence and thus is not used as context. A \textit{visible} token has a concrete prediction and can participate in subsequent denoising steps, enabling decoding to progress in parallel. However, visible tokens are not yet treated as complete: their token identities and KV states may still be revised as the surrounding context becomes more complete. A \textit{stable} token has passed a stronger stability criterion and is considered reliable enough for long-term reuse.

\name{} uses two thresholds to control token state transitions. At each denoising step, tokens that satisfy the \textit{visibility threshold} $\theta_{\text{visible}}$ are promoted to the visible state, while tokens that further satisfy the \textit{stability threshold} $\theta_{\text{stable}}$ are promoted to the stable state. Both visible and stable tokens can be exposed for parallel decoding, but only stable tokens are permanently reused as prefix context. If no token satisfies the $\theta_{\text{visible}}$, \name{} promotes the most confident token to visible to ensure continuous progress. Importantly, early visibility does not imply permanent commitment: visible tokens remain tracked, even after their blocks have been reused as prefix context, until they satisfy the $\theta_{\text{stable}}$ and are safely merged into the long-term prefix KV cache.

\noindent\textbf{Sparse Cache Refresh.} The next challenge is to efficiently track and refresh visible tokens once they enter the prefix context. These revisions are sparse, irregular, and position-dependent, often involving only a few tokens scattered across previous blocks. Since such computation poorly matches the NPU dense execution model, performing it on the NPU would disrupt current-block denoising and add data movement and synchronization overheads.

To avoid this inefficiency, \name{} separates sparse token maintenance from dense NPU decoding. As shown in Figure~\ref{fig:method-revision}(b), \name{} introduces a CPU-side path that asynchronously recomputes token confidence, refreshes token identities, and updates their KV states. This division matches the hardware characteristics of the two processors: the NPU handles high-throughput dense computation, while the CPU handles sparse updates with lower scheduling overhead.

\name{} enables the CPU path with two techniques. First, the CPU refresh path reuses the same HMX-specialized weights as the main NPU model. Because these weights are stored in NPU-specific tiled and quantized layouts, standard CPU matrix multiplication cannot directly interpret them under the original row-column semantics. \name{} therefore implements specialized CPU kernels that explicitly decode the HMX tile permutation and quantized storage order, ensuring that CPU-side recomputation preserves the shared memory layout and avoids inconsistent computation.

Second, \name{} adopts a delayed-merge policy to keep CPU refreshes off the critical path. At the beginning of each denoising step, the NPU reads the current prefix KV cache and proceeds with current-block decoding. In parallel, the CPU refreshes selected KV entries based on the latest token states. These refreshed entries are not consumed by the NPU within the same step; instead, they are merged back into the prefix KV cache only at step boundaries. This step-boundary merge avoids fine-grained CPU-NPU synchronization while making refreshed context available to later denoising steps.

\noindent\textbf{Selective Logits Skipping.} In addition to sparse cache refreshes, \name{} offloads logits computation to the CPU path to relieve NPU memory pressure from vocabulary-sized output projection weights. Since naively computing logits for all tokens on the CPU would incur substantial overhead, \name{} performs selective logits skipping to keep CPU computation manageable. The key observation is that logits are needed only for positions whose token identities may still change or whose confidence must be evaluated for future decisions. For stable tokens whose identities no longer affect commit decisions, \name{} skips output projection, confidence estimation, and sampling logic.

\subsection{Swap-Optimized Memory Runtime}
\begin{figure}[t]
    \centering
    \includegraphics[width=1.0\linewidth]{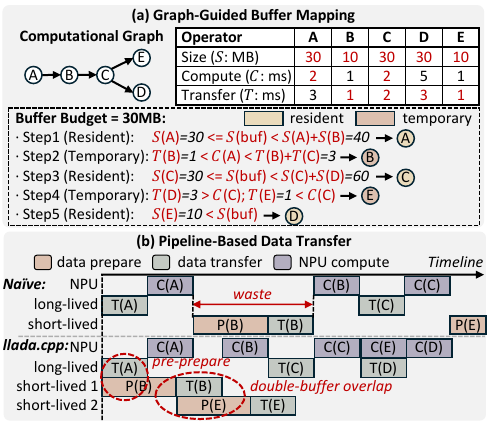}
    \caption{(a) \name{} uses the computational graph and profiled metrics to determine which tensors remain resident in NPU-visible memory. (b) \name{} uses double buffering to hide data transfer latency for short-lived tensors.}
    \label{fig:method-runtime}
\end{figure}

Mobile NPUs access system memory through specific buffers (e.g., RPCMEM~\cite{qcom-fastrpc} or DMABUF~\cite{linux-dmabuf}), which must be mapped into the NPU virtual address (VA) space before kernel execution. However, this space is limited and often cannot simultaneously hold model weights, the KV cache, and workspaces, forcing frequent data swapping with extra mapping and transfer overheads. As shown in Figure~\ref{fig:method-runtime}, \name{} builds a swap-optimized memory runtime that minimizes avoidable VA swaps and makes unavoidable swaps more efficient by hiding their latency behind NPU execution.

\noindent\textbf{Graph-Guided Buffer Mapping.} \name{} first reduces unnecessary VA swaps by organizing buffer mappings according to the computational graph, which provides the execution order of NPU operators and the producer-consumer relationships of tensors. Using this information, the runtime classifies tensors as either \textit{long-lived} or \textit{short-lived} under the available VA budget. Tensors that must remain visible across denoising steps, such as the model weights and KV cache, are treated as long-lived by default. In contrast, activations and temporary tensors are classified as short-lived because their live ranges are bounded by local graph dependencies.

When the VA budget cannot keep all long-lived data resident, \name{} uses a profile-guided placement algorithm to determine which objects remain resident and which are mapped temporarily. Figure~\ref{fig:method-runtime}(a) illustrates this procedure. During initialization, \name{} profiles representative operators to collect three types of information: (1) the size of associated long-lived tensors, (2) NPU execution time, and (3) the transfer time required to prepare inputs on the host and make them visible in the NPU VA space. Using this profile, \name{} constructs alternating \textit{resident data} and \textit{temporary data}. Guided by the execution order, \name{} traverses all operators and classifies their associated tensors. Given the buffer budget, \name{} first fills it with as much resident data as possible and then selects temporary data whose transfers can be maximally overlapped with computation on resident data. In particular, dependency-free operators can be reordered to better satisfy the overlap criterion.

For long-lived data that remain resident, \name{} assigns stable VA regions that stay mapped across denoising steps. These tensors are further packed according to their graph-derived consumption order. Particularly, tensors used by consecutive NPU operators are placed close to each other and accessed through a small number of large buffer mappings. For short-lived data, including naturally temporary tensors and long-lived data demoted by the classification algorithm, \name{} uses compact scratch or staging regions and reuses offsets based on graph-derived lifetimes. Tensors with non-overlapping live ranges share the same VA range, while operator-local workspaces are released immediately after use. This organization consolidates fragmented tensors into fewer mappings, reduces avoidable VA swaps, and lowers the bookkeeping cost of VA management.

\noindent\textbf{Pipeline-Based Data Transfer.} After compacting buffer mappings, \name{} reduces unavoidable VA-swap overhead by prefetching mapping metadata and staging data. Guided by the computational graph, the runtime prepares buffer handles, offsets, sizes, and target staging slots while the NPU executes preceding operators. For short-lived data that must be staged before use, especially long-lived data demoted by the placement algorithm, this prefetch step also identifies the target staging slot and starts preparing the data before its consumer operator is reached. This preserves tensor layout and computation semantics while moving mapping preparation and data staging off the critical path.

\name{} uses double-buffered staging to hide transfer latency. The runtime allocates two reusable staging slots within the NPU-visible VA space. While the NPU consumes data from one slot, the host prepares the next required data object in the other slot by preparing its buffer descriptor and validating its VA mapping. Once the current operator finishes, the two slots exchange roles: the prepared slot becomes the input to the next NPU kernel, and the released slot is reused for the following transfer. Following the graph-derived access sequence, the runtime prefetches data ahead of its consumer operators, overlapping transfer and mapping costs with resident NPU computation whenever possible.

\section{Implementation}
We implement \name{} as an end-to-end framework built on \textit{llama.cpp}~\cite{llama-cpp}, extending it with over 12K lines of code. Our implementation targets Qualcomm Hexagon NPUs due to their relatively open software ecosystem, while the core design can be adapted to other mobile platforms with similar execution paradigms. \name{} consists of two modules:

The first module is a \textit{Qualcomm Hexagon NPU operator library}, compiled as a Hexagon DSP shared object using the Hexagon SDK~\cite{qcom-hexagon-sdk}. It targets the DSP-coupled HMX~\cite{qcom-hmx} and HVX~\cite{qcom-hvx} engines, and implements FP16 matrix multiplication, Q4\_0 matrix multiplication for 4-bit quantized weights, FlashAttention over FP16 KV states, RMSNorm, elementwise operators, shared-memory mapping, and a device-side worker pool. The model converter extends the GGUF pipeline~\cite{gguf} with a layout pass that rearranges FP16 weights into 32-by-32 HMX tiles and packs quantized weights for HVX-based dequantization kernels.

The second module is an \textit{Android host runtime}, which adds a GGML Hexagon backend~\cite{ggml}, a dLLM model path, and a block-wise diffusion driver. It invokes the device library through FastRPC and uses RPCMEM shared memory~\cite{qcom-fastrpc}, backed by Linux DMABUF~\cite{linux-dmabuf}, as the tensor buffer type, allowing the CPU and NPU to access the same physical pages without copies. Before dispatching each operator, the backend validates NPU-visible mappings and passes tensors as file-descriptor/offset pairs. Supported operators are executed on the NPU, while the host manages block-wise denoising, KV cache views, token-state transitions, refresh candidates, lookahead positions, and output decisions.

\section{Evaluation}
\subsection{Experimental Setup}
\noindent\textbf{Platform.} We evaluate \name{} on three Qualcomm smartphones (Table~\ref{tab:exp_env}). The OnePlus Ace5 Pro with the Snapdragon 8 Elite (SM8750) SoC is the primary platform used for all end-to-end, ablation, and energy experiments. The OnePlus 12 with the Snapdragon 8 Gen3 (SM8650) and the OnePlus 15 with the Snapdragon 8 Elite Gen5 (SM8850) are used in the cross-device study to test portability across SoC generations. They represent three successive SoC generations with progressively higher NPU compute capability. All three phones run Android and execute the same \name{} binary on the Hexagon NPU. We enable the NPU FlashAttention~\cite{flashattention} operator on SM8650 and SM8750, while SM8850 uses the CPU FlashAttention operator due to runtime support limitations.

\begin{table}[t]
\centering
\caption{Smartphone configurations.}\label{tab:exp_env}
\small
\setlength{\tabcolsep}{2.5pt}
\begin{tabular*}{\columnwidth}{@{}l@{\extracolsep{\fill}}ccc@{}}
    \toprule
    \textbf{Device} & \textbf{SoC} & \textbf{Memory} & \textbf{NPU INT8 TOPS} \\
    \midrule
    OnePlus 12 & SM8650 & 16 GB LPDDR5X & 45~\cite{gadgetversus-sm8650} \\
    OnePlus Ace5 Pro & SM8750 & 16 GB LPDDR5X & 65.25~\cite{gadgetversus-sm8750} \\
    OnePlus 15 & SM8850 & 16 GB LPDDR5X & 89.4~\cite{gadgetversus-sm8850} \\
    \bottomrule
\end{tabular*}
\end{table}

\noindent\textbf{Models.} LLaDA-8B-Instruct~\cite{llada} is the primary workload, as it is a mainstream dLLM model in recent studies. We also include Llama-3-8B-Instruct~\cite{llama3} as an autoregressive reference in the end-to-end generation, quality, and energy studies. All models use Q4\_0 low-bit weight quantization to enable LLaDA and Llama inference on smartphones. In addition, we adapt Dream-7B~\cite{dream}, another dLLM whose most notable architectural difference from LLaDA is its grouped-query attention with fewer KV heads, which changes attention computation. This cross-model experiment tests whether \name{} benefits extend beyond one dLLM architecture.

\noindent\textbf{Tasks.} We evaluate generation quality on 200-sample subsets from each of four datasets (Table~\ref{tab:eval_tasks}). These tasks are used both for model-level comparison between LLaDA and Llama and for isolating the quality impact of \name{} components.

\begin{table}[t]
    \centering
    \caption{Evaluation datasets and task types.}
    \label{tab:eval_tasks}
    \small
    \setlength{\tabcolsep}{4pt}
    \begin{tabular*}{\columnwidth}{@{}l@{\extracolsep{\fill}}p{0.68\columnwidth}@{}}
    \toprule
    \textbf{Dataset} & \textbf{Task} \\
    \midrule
    GSM8K~\cite{gsm8k} & Grade-school math problem dataset \\
    BoolQ~\cite{boolq} & Boolean question answering dataset \\
    ARC-C~\cite{arc} & Science question answering dataset \\
    HellaSwag~\cite{hellaswag} & Commonsense completion dataset \\
    \bottomrule
    \end{tabular*}
\end{table}

\noindent\textbf{Baselines and settings.} We compare \name{} with three baselines. The \emph{CPU baseline} follows vanilla dLLM decoding and repeatedly denoises the full sequence on the smartphone CPU. The \emph{CPU with prefix KV cache reuse} baseline keeps computation on the CPU but applies prefix KV cache reuse across completed blocks. The \emph{NPU baseline} keeps the vanilla denoising schedule but offloads dense Transformer computation, mainly large matrix multiplications, to the NPU. Their denoising step count is set to the number of target generated tokens by default. Unless otherwise stated, dual-path progressive revision uses visibility and stability thresholds of 0.7 and 0.9. For performance breakdown and sensitivity analysis, our implementation can enable or disable prefix KV cache reuse, NPU offloading, and each proposed component independently while keeping the other settings fixed.

\noindent\textbf{Metrics.} We report end-to-end latency, generation quality, average power, and energy per request. Latency is measured as wall-clock request time, including prompt processing and generation. For each dataset, we measure five cases, repeat each run three times, and report exact generated-token latency for 32, 64, or 128 target tokens. Since the default dLLM block size is 32 tokens, this range covers both single-block and multi-block generation cases. All generations run with temperature 0. Generation quality follows the task-specific scoring logic in lm-evaluation-harness~\cite{lm-eval-harness}. We map each model output to the corresponding task answer format and report accuracy. Power is sampled from battery sysfs traces~\cite{sysfs}, and energy per request is computed by integrating idle-subtracted power over the request window.

\subsection{End-to-End dLLM Generation}

\begin{figure*}[t]
    \centering
    \includegraphics[width=\textwidth]{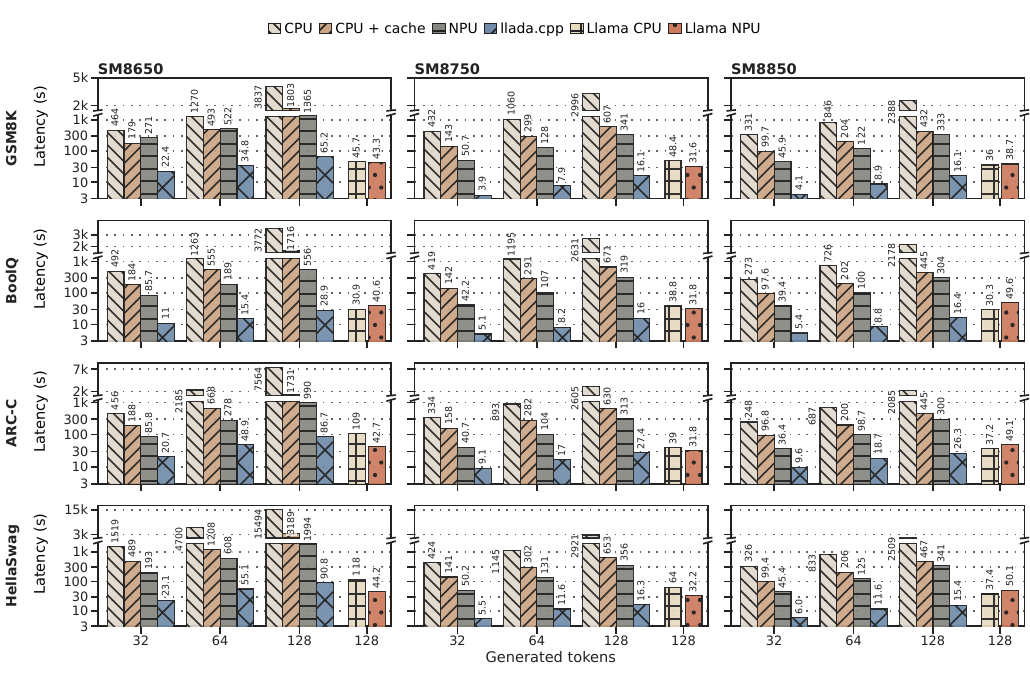}
    \caption{End-to-end generation latency on different smartphones. LLaDA groups report 32-, 64-, and 128-token outputs under CPU, CPU with prefix KV cache reuse, NPU, and full \name{} execution; Llama is a 128-token autoregressive reference.}
    \label{fig:e2e-performance}
\end{figure*}

Figure~\ref{fig:e2e-performance} compares the time needed to generate 32, 64, and 128 tokens. We report latency across four datasets because dLLM decoding speed depends on token predictability, and different tasks can produce different confidence patterns and effective denoising steps. This effect is evident under the same setup, and the end-to-end latency differs noticeably across datasets. As NPU compute capability increases from SM8650 to newer SoCs, the absolute latency of \name{} generally decreases. However, the relative speedup does not grow monotonically, because the baselines also benefit from newer platforms and the optimized execution starts to expose scheduling, token-state management, and CPU-side work.

The comparison also shows that only NPU offloading is insufficient. The NPU baseline accelerates dense Transformer computation, but it still follows the vanilla denoising schedule and repeatedly pays for full sequence forwards. As a result, it does not separate clearly from the CPU baseline on some settings. \name{} reduces this gap by increasing useful token work in each NPU forward pass and avoiding unnecessary output decisions for stable tokens, making NPU-backed dLLM generation practical on recent SoCs. For 128-token generation, \name{} is still slower than Llama on SM8650, where NPU compute is more limited, but becomes faster than Llama on SM8750 and SM8850. The result of Llama shows a smaller and less consistent NPU benefit, because autoregressive decoding exposes much less token-level parallelism to the NPU than block-wise dLLM decoding.

\begin{figure}[t]
    \centering
    \includegraphics[width=\columnwidth]{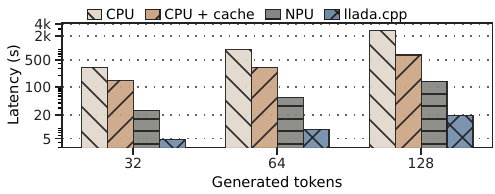}
    \caption{End-to-end generation latency on Dream-7B with different generated-token lengths, comparing CPU, CPU with prefix KV cache reuse, NPU, and \name{}.}
    \label{fig:dream7b-cross-model}
    \vspace{-0.5cm}
\end{figure}

We adapt Dream-7B and measure its performance on the SM8750 device. Figure~\ref{fig:dream7b-cross-model} shows acceleration trends similar to LLaDA-8B, demonstrating the scalability of our method.

\subsection{Quality Preservation}
Table~\ref{tab:quality} first compares LLaDA with the Llama autoregressive reference. LLaDA remains competitive across the four tasks, which supports using it as a usable dLLM on smartphones. The lower block then isolates the quality impact of \name{} components. Moving dense computation to the NPU does not introduce a systematic accuracy drop, while multi-block speculative decoding alone can reduce accuracy because future-block work is used before the surrounding context is fully stable. Dual-path progressive revision brings quality back close to the CPU path by keeping early visible tokens revisable and refreshing unstable prefix states.

\begin{table}[t]
    \centering
    \caption{Generation quality on four 200-sample datasets. The upper block compares LLaDA and Llama under CPU and NPU-backed settings, while the lower block reports the LLaDA component ablation with multi-block speculative decoding (MBSD) and dual-path progressive revision (DPPR).}
    \label{tab:quality}
    \small
    \setlength{\tabcolsep}{3pt}
    \begin{tabular*}{\columnwidth}{l@{\extracolsep{\fill}}rrrr}
    \toprule
    \textbf{Setting} & \textbf{GSM8K} & \textbf{BoolQ} & \textbf{ARC-C} & \textbf{HellaSwag} \\ \midrule
    \multicolumn{5}{l}{\textit{Model-level quality}} \\
    LLaDA CPU & 39.0 & 82.5 & 84.0 & 51.0 \\
    LLaDA \name{} & 43.5 & 80.5 & 85.0 & 49.5 \\
    Llama CPU & 30.0 & 83.5 & 71.0 & 67.0 \\
    Llama NPU & 31.0 & 77.5 & 78.5 & 60.0 \\
    \midrule
    \multicolumn{5}{l}{\textit{LLaDA component ablation}} \\
    CPU & 39.0 & 82.5 & 84.0 & 51.0 \\
    NPU & 41.5 & 82.0 & 83.5 & 50.0 \\
    NPU + MBSD & 37.5 & 79.5 & 82.0 & 46.0 \\
    NPU + MBSD + DPPR & 43.5 & 80.5 & 85.0 & 49.5 \\
    \bottomrule
    \end{tabular*}
\end{table}

\begin{table}[t]
    \centering
    \caption{128-token generation performance breakdown of \name{} using GSM8K on SM8750, where the rows gradually add prefix KV cache reuse, swap-optimized memory runtime (SOMR), multi-block speculative decoding (MBSD), staged token stabilization (STS), selective logits skipping (SLS), and sparse cache refresh (SCR).}
    \label{tab:performance-breakdown}
    \small
    \setlength{\tabcolsep}{3.0pt}
    \begin{tabular*}{\columnwidth}{l@{\extracolsep{\fill}}rrr}
    \toprule
    \textbf{Method} & \textbf{Latency} & \textbf{Rel. Gain} & \textbf{Tot. Gain} \\ \midrule
    CPU & 2996.2~s & 1.00$\times$ & 1.00$\times$ \\
    CPU + prefix KV cache & 607.0~s & 4.94$\times$ & 4.94$\times$ \\
    NPU & 341.2~s & 1.78$\times$ & 8.78$\times$ \\
    NPU + prefix KV cache & 87.0~s & 3.92$\times$ & 34.44$\times$ \\
    + SOMR & 71.6~s & 1.22$\times$ & 41.87$\times$ \\
    + MBSD & 59.8~s & 1.20$\times$ & 50.08$\times$ \\
    + STS & 16.4~s & 3.65$\times$ & 182.55$\times$ \\
    + SLS & 15.8~s & 1.04$\times$ & 189.38$\times$ \\
    + SCR & 16.1~s & 0.98$\times$ & 186.34$\times$ \\
    \bottomrule
    \end{tabular*}
\end{table}

\subsection{Performance Breakdown}\label{subsec:performance-breakdown}
Table~\ref{tab:performance-breakdown} reports the 128-token latency breakdown on the SM8750 device using the GSM8K cases as the main latency experiment. Prefix KV cache reuse reduces repeated prefix computation in both CPU and NPU baselines, while moving dense denoising to the NPU provides a larger system-level reduction. Swap-optimized memory runtime further lowers latency by reducing recurrent NPU buffer remapping, but its isolated gain is smaller because dense denoising forward passes still dominate the NPU baseline. The largest additional reductions come from the decoding components. Multi-block speculative decoding fills late-stage forwards with future-block work, and Staged Token Stabilization accelerates decoding by separating early visibility from long-term stability. Selective Logits Skipping removes redundant output decisions for stable tokens. Its benefit varies by input, since early denoising steps may stabilize many tokens while later steps only decide a few tokens, allowing stable-position logits to be skipped and reducing latency by more than 10\% in some cases. Sparse Cache Refresh adds the quality-preserving CPU-side refresh step with small latency overhead.

\begin{figure}[t]
    \centering
    \includegraphics[width=\columnwidth]{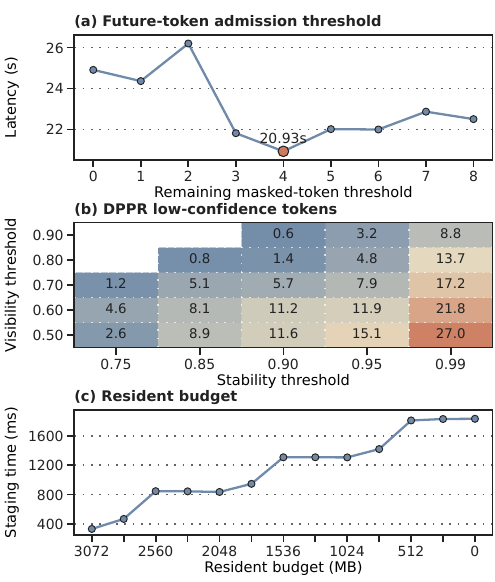}
    \caption{Sensitivity analysis for \name{} components. (a) Latency as future-block tokens are admitted at different remaining masked-token thresholds in multi-block speculative decoding (MBSD). (b) Tokens that remain below the stability threshold when the current block finishes, under different dual-path progressive revision (DPPR) visibility and stability thresholds. (c) Synchronous staging time under different NPU-visible weight budgets in swap-optimized memory runtime (SOMR).}
    \label{fig:sensitivity}
\end{figure}

\subsection{Sensitivity Analysis}\label{subsec:sensitivity}
Figure~\ref{fig:sensitivity} focuses on the runtime-level control parameters that directly correspond to \name{} components. Figure~\ref{fig:sensitivity}(a) shows that the threshold for admitting future-block tokens has a clear operating point on the device. Triggering at four remaining tokens gives the lowest latency, while triggering too early adds unnecessary future-block work and triggering too late misses useful NPU work. Figure~\ref{fig:sensitivity}(b) reports how many tokens still fail the stability threshold when the current block finishes under different visibility and stability thresholds. This result guides the threshold choice in dual-path progressive revision. A lower visibility threshold exposes more tokens earlier and improves decoding progress, but it also leaves more visible tokens that may still need revision. A higher stability threshold protects quality by keeping uncertain tokens out of the long-term prefix KV cache, but it increases the amount of additional CPU-side refresh work. Figure~\ref{fig:sensitivity}(c) fixes the temporary staging budget at 256~MB and reduces the NPU-visible weight budget. The model remains runnable even when no weights are kept visible across steps, but synchronous staging time increases as more weights must be streamed through staging buffers.

\subsection{Memory Boundary and Long-Output Runs}
\noindent\textbf{Memory boundary.} Repeated denoising revisits the same weights, prefix KV states, activations, and temporary buffers across many steps. This makes NPU-visible address-space management part of the dLLM critical path rather than a one-time initialization cost. Table~\ref{tab:memory-boundary} reports the measured boundary runs under the same SM8750 device configuration. Without swap-optimized memory runtime, the NPU baseline fails before producing tokens at microbatch sizes 256 and 512. With graph-guided buffer mapping and pipelined staging, \name{} completes both boundary settings.

\noindent\textbf{Long-output runs.} A longer generation makes the prefix KV cache grow and keeps more model states active across denoising steps, which increases pressure on the NPU-visible address space. Swap-optimized memory runtime keeps these larger execution points runnable by combining graph-guided buffer mapping with pipelined staging. This allows \name{} to support longer-context dLLM generation on smartphones, improving its practical usability beyond short requests.

\begin{table}[t]
    \centering
    \caption{Memory-boundary runs on SM8750. The table varies the maximum generated-token budget and compares the full swap-optimized memory runtime (SOMR) with a paging-only setting that keeps replacement but disables graph-guided buffer mapping and pipelined staging.}
    \label{tab:memory-boundary}
    \small
    \setlength{\tabcolsep}{3pt}
    \begin{tabular*}{\columnwidth}{l@{\extracolsep{\fill}}rrl}
    \toprule
    \textbf{Method} & \textbf{Token Budget} & \textbf{Target} & \textbf{Output} \\
    \midrule
    No SOMR & 256 & 256 & failed \\
    \name{} with SOMR & 256 & 256 & 224 \\
    No SOMR & 512 & 512 & failed  \\
    Paging only & 512 & 512 & failed \\
    \name{} with SOMR & 512 & 512 & 401 \\
    \bottomrule
    \end{tabular*}
\end{table}

\subsection{Energy Behavior}
\begin{figure}[t]
    \centering
    \includegraphics[width=\linewidth]{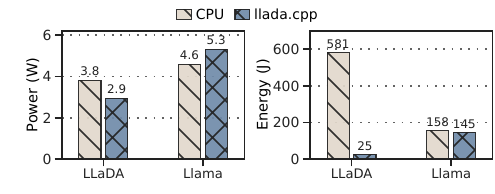}
    \caption{Comparison of average power and energy consumption on SM8750 under fixed 128-token workloads.}
    \label{fig:energy-thermal}
\end{figure}

We measure battery power on SM8750 using a fixed 128-token workload. Figure~\ref{fig:energy-thermal} reports average power and idle-subtracted energy. For LLaDA, \name{} reduces both power and per-request energy by giving the NPU sufficient parallel work to finish quickly. In contrast, NPU-accelerated Llama shows higher average power than its CPU setting: autoregressive decoding repeatedly issues small NPU workloads, incurring accelerator scheduling overhead without enough token-level parallelism to amortize it. As a result, its limited latency reduction largely cancels the energy benefit. This contrast shows that mobile NPU energy efficiency depends on exposing sufficient parallel token work, not merely on offloading an LLM to the NPU.

\section{Related Works}
\noindent\textbf{On-Device LLM Inference.} Recent work has explored quantization, heterogeneous execution, and mobile NPU specialization for LLM inference. \textit{llm.npu}~\cite{llmnpu} maps autoregressive LLMs to NPUs using INT8 NPU kernels, CPU-side outlier compensation, chunk-sharing graphs, and out-of-order subgraph scheduling. \textit{llama.cpp-npu}~\cite{hao2025npu} reverse-engineers Hexagon HMX execution and builds a mobile-NPU path for test-time scaling with hardware-aware tile quantization and LUT-based vector kernels. HeteroLLM~\cite{heterollm} characterizes mobile SoC execution and improves heterogeneous LLM inference through GPU-NPU tensor partitioning, fast synchronization, and reusable memory pools. These systems show that mobile acceleration depends on layout, scheduling, synchronization, and memory organization, rather than peak TOPS alone. However, they primarily target autoregressive decoding or static dense inference graphs, where each decoding step exposes limited matrix parallelism on the NPU. In contrast, \name{} targets block-wise dLLM inference, whose parallel denoising naturally forms dense NPU forwards and converts into lower latency.

\noindent\textbf{dLLM Inference Optimization.} Prior work~\cite{d3pm,llada,mercury} reduces the cost of iterative denoising over masked sequences. Block-wise decoding~\cite{block-diffusion} makes long generation practical by combining intra-block parallel updates with left-to-right block progression. Fast-dLLM~\cite{fast-dllm} enables KV cache reuse and confidence-aware parallel decoding for dLLMs, while cache-oriented methods such as dKV-Cache~\cite{dkv-cache}, dLLM-Cache~\cite{dllm-cache}, and FlashBlock~\cite{flashblock} exploit step-to-step stability in KV states or attention outputs. Other techniques improve decoding through remasking or token editing~\cite{remdm,llada21}, distillation and parallel decoding~\cite{d3llm,dparallel}, or early skipping~\cite{es-dllm}. These methods primarily optimize decoding algorithms or reusable model states. \name{} focuses on algorithm-system co-design, aligning the parallel decoding nature of dLLMs with the tensor-computation capabilities of mobile NPUs.

\section{Conclusion}
We presented \name{}, the first NPU-aware inference framework for accelerating dLLMs on smartphones. We hope \name{} paves the way for low-latency, privacy-preserving, and widely accessible AI capabilities on mobile devices.

\bibliographystyle{ACM-Reference-Format}
\bibliography{reference}

\end{document}